\documentclass{article}
\usepackage{ijcai25}

\usepackage{times}
\usepackage{float}
\usepackage{soul}
\usepackage{makecell}
\usepackage{url}
\usepackage{authblk}
\usepackage[hidelinks]{hyperref}
\usepackage[utf8]{inputenc}
\usepackage[small]{caption}
\usepackage{graphicx}
\usepackage{tablefootnote}
\usepackage{amsmath}
\usepackage{amsthm}
\usepackage{amssymb}
\usepackage{hyperref}
\usepackage{mathrsfs} 
\usepackage{booktabs}
\usepackage{algorithm}
\usepackage{algorithmic}
\usepackage{color}
\usepackage{multirow}
\usepackage{hyperref}
\usepackage{subcaption}
\usepackage[switch]{lineno}

\title{STAMImputer: Spatio-Temporal Attention MoE for Traffic Data Imputation}

\newcommand{\corresymbol}{*}

\author[1]{\textbf{Yiming Wang}}
\author[1, 2, 3]{\textbf{Hao Peng}\textsuperscript{\corresymbol}}
\author[4]{\textbf{Senzhang Wang}\textsuperscript{\corresymbol}}
\author[1]{\textbf{Haohua Du}\textsuperscript{\corresymbol}}
\author[5]{\textbf{Chunyang Liu}}
\author[6]{\textbf{Jia Wu}}
\author[7]{\textbf{Guanlin Wu}\textsuperscript{\corresymbol}}

\affil[1]{School of Cyber Science and Technology, Beihang University}
\affil[2]{Hangzhou Innovation Institute of BUAA, Hangzhou, China}
\affil[3]{Department of Computer Science and Technology, Shantou University}
\affil[4]{School of Computer Science and Engineering, Central South University}
\affil[5]{Didi Chuxing}
\affil[6]{Department of Computing, Macquarie University}
\affil[7]{National University of Defense Technology, Changsha, China}
\affil[ ]{\text{\{yiimngwang, penghao, duhaohua\}@buaa.edu.cn, szwang@csu.edu.cn,}}
\affil[ ]{\text{liuchunyang@didiglobal.com, jia.wu@mq.edu.au, wuguanlin16@nudt.edu.cn.}}

\begin{document}

\maketitle
\begin{abstract}
    Traffic data imputation is fundamentally important to support various applications in intelligent transportation systems such as traffic flow prediction.
    However, existing time-to-space sequential methods often fail to effectively extract features in block-wise missing data scenarios. 
    Meanwhile, the static graph structure for spatial feature propagation significantly constrains the model's flexibility in handling the distribution shift issue for the nonstationary traffic data.
    To address these issues, this paper proposes a 
    \textbf{S}patio\textbf{T}emporal \textbf{A}ttention \textbf{M}ixture of experts network named \textbf{STAMImputer} for traffic data imputation.
    Specifically, we introduce a Mixture of Experts (MoE) framework to capture latent spatio-temporal features and their influence weights, effectively imputing block missing.
    A novel \textbf{L}ow-\textbf{r}ank guided \textbf{S}ampling \textbf{G}raph \textbf{AT}tention (\textbf{LrSGAT}) mechanism is designed to dynamically balance the local and global correlations across road networks.
    The sampled attention vectors are utilized to generate dynamic graphs that capture real-time spatial correlations.
    Extensive experiments are conducted on four traffic datasets for evaluation. The result shows STAMImputer achieves significantly performance improvement compared with existing SOTA approaches.
    Our codes are available at~\url{https://github.com/RingBDStack/STAMImupter}.
\end{abstract}
\renewcommand{\thefootnote}{\fnsymbol{footnote}}
\footnotetext[1]{Corresponding authors.}
\section{Introduction}
\label{sec_Intro}
Intelligent Transportation Systems (ITS), which rely on comprehensive and high-quality data to perform their tasks effectively~\cite{liu2023cross,yin2021deep}, play a significant role in maintaining urban traffic order.
However, in real-world traffic data collection, limitations in resources such as sensors (e.g., vehicles, drones) and disruptions in data acquisition (e.g., network interruptions, extreme weather conditions) can lead to data missing in both spatio-temporal dimensions~\cite{chan2023missing,zhang2024comprehensive}.
Therefore, traffic data imputation, which aims to reconstruct the missing traffic data by leveraging potential data correlations, has attracted rising research attention recently.

Existing research on traffic data imputation can be broadly categorized into time-series imputation and spatio-temporal imputation.
Time-series imputation primarily involves algorithms based on statistical analysis~\cite{yu2016temporal,chen2020nonconvex,chen2021bayesian} or deep learning models that concentrate efforts exclusively on the temporal dimension of the traffic data~\cite{cao2018brits,du2023saits,zhang2024score}.
However, these methods are not effective in addressing the issue of block data missing, where data is absent in consecutive sequences.
Spatio-temporal imputation~\cite{xu2022traffic,wang2022generative,li2022fine,nie2024imputeformer} accounts for the extra spatial dimension with static graphs or spatial attentions, allowing for more effective handling of block missing data.

Although considerable effort has been made on traffic data imputation, there are still two major challenges that hinder existing works from achieving a more accurate imputation result.
First, most current time-to-space sequential learning frameworks~\cite{marisca2022learning,wei2024self,zou2024multispans}, which process spatio-temporal data by first using a temporal module followed by a spatial module, may extract and propagate invalid features when block data are missing.
The block data missing scenario can occur in both spatial and temporal dimensions. 
This challenge is even more pronounced in scenarios with spatio-temporal block-wise missing data, for example, due to prolonged power outages in physical sensors. 
Second, static graph-based methods struggle to capture global spatial dependencies beyond local communities, which are essential for making accurate predictions in practical applications. 
For example, the traffic volume or average traffic speed of a crossroad during non-peak hours is closely related to its adjacent intersections. 
However, during the evening peak, these observations are also affected by nearby traffic hub nodes, which are key sensor locations that have the most significant impact on the entire traffic network.
Many static graph-based methods~\cite{ye2021spatial,xu2024hierarchical} are limited in their ability to capture the global spatial correlations.
Without static graphs, global spatial attention methods~\cite{zhang2022self,nie2024imputeformer} can propagate spatial features across traffic networks but struggle to capture spatial correlations in highly sparse data due to the lack of guidance from prior knowledge.
Hence, we argue that spatio-temporal imputation requires dynamically adjustment according to the real-time missing data.
Furthermore, combining the strengths of static graph-based and global attention-based methods will effectively handle complex imputation scenarios.

To address the above challenges, we propose a \textbf{S}patio-\textbf{T}emporal self-\textbf{A}ttention \textbf{M}ixture of experts \textbf{Imputer} network named STAMImputer.
We perform traffic data imputation by deploying specialized experts networks, including attention experts and observation experts.
Attention experts are self-attention transformers that capture temporal or spatial features, while observation experts act as arbitrators, overseeing and controlling the broad spatio-temporal attention.
The MoE framework dynamically adjusts real-time spatio-temporal attention weights by utilizing the routing control of the observation experts.
If the data is sparse in terms of the spatial demension but rich in terms of temporal dimension, the observation experts guide the model to prioritize the temporal experts' learned representations.
Moreover, inspired by ~\cite{fang2023spatio} and~\cite{nie2024imputeformer}, we design a novel \textbf{L}ow-\textbf{r}ank guided \textbf{S}ampling \textbf{G}raph \textbf{AT}tention mechanism, namely \textbf{LrSGAT}. 
The spatial low-rank guided matrix decoupling identifies key features of traffic hubs while filtering out the redundant global relationships. 
Attention projection is then applied to propagate the relevant features of these hubs.
Finally, the sampled attention vectors contributes to generating dynamic graphs which fully capture the real-time spatial correlations and help improve downstream traffic-related tasks. 
Our contributions are summarized as follows.

$\bullet$ We propose a novel traffic data imputation model based on the MoE framework.
Our model dynamically adjusts attention weights of spatio-temporal expert networks to balance the contributions of both spatio-temporal dimensions.
To the best of our knowledge, this is the first application of the MoE framework to traffic data imputation tasks.

$\bullet$ We propose a low-rank guided sampling graph attention mechanism that enables the imputation model to capture global spatial dependencies beyond local communities.

$\bullet$ We build real-time dynamic graphs by utilizing the attention vectors sampled by the spatial expert network, which accurately represent the current traffic conditions.

$\bullet$ Extensive comparative experiments on four real-world benchmark datasets demonstrate that our STAMImputer model outperforms others in traffic data imputation.
\section{Related Works}
\label{sec_Rel}

Traditionally, statistical based approaches~\cite{nelwamondo2007missing,van2011mice,yi2016st} are used to study the mathematical characteristics and functional properties of spatio-temporal sparse data, such as low-rank approximation~\cite{cichocki2009fast}. 
In terms of spatio-temporal dual-dimensions, ~\cite{li2013efficient} proposed an algorithm based on spatio-temporal probabilistic principal component analysis (PPCA).
Both \cite{deng2021graph} and \cite{xu2023hrst} proposed low-rank guided tensor completion algorithms. 
The former used graph spectral regularization, while the latter introduced a Hessian regularization spatio-temporal low-rank method.

More recently, deep learning approaches have been proposed for traffic data imputation.
Early works include Autoencoder, RNN and CNN networks~\cite{cao2018brits,benkraouda2020traffic,zhao2020traffic}, and they rely on a keen sense of temporal variation to impute the missing data. 
As development progressed, graph-based methods are introduced~\cite{chen2022novel,liang2022memory,kong2023dynamic}, contributing spatial dimensionality correlations to the imputation work.
Notably, \cite{cini2021filling} proposed bidirectional autoregressive architecture which conducted spatio-temporal imputation through message passing graph neural networks.
Nevertheless, GANs~\cite{yoon2018gain,chen2019traffic} are also used in traffic data imputation.
For example, STGAN~\cite{yuan2022stgan} implemented a spatio-temporal GAN model to perceive local and global spatio-temporal distribution.

In more recent work, the self-attention mechanism provides a fresh and excellent choice for imputation work in spatio-temporal dimensions~\cite{vaswani2017attention}.
\cite{ma2019cdsa} proposed a cross-dimensional attention completion method, which, for the first time, applied the self-attention mechanism to multivariate geotagged time series data to achieve a cross-dimensional joint capture of self-attention.
The SPIN model in~\cite{marisca2022learning} implemented an spatio-temporal attention based framework, whitch can handle sparse data without propagating prediction errors or requiring a bidirectional model to encode forward and backward time dependencies.
\cite{nie2024imputeformer} adopted attention on both spatio-temporal dimensions.
Combining low-rank-induced temporal attention and spatial embedding attention, it demonstrates state-of-the-art performance in traffic spatio-temporal data imputation.
However, we argue that none of the above-cited methods can dynamically balance the spatio-temporal dual-dimensional learning features.
\section{Problem Definition}
\label{sec_Prob}
The traffic completion problem aims to infer unobserved or unrecorded data using known traffic data recorded by sensors.
Specifically, $\mathcal{X}_{t:t+T} \in \mathbb{R}^{T \times N \times C}$, where $C=1$, represents the observed traffic flow or speed value of $N$ nodes in traffic network from a fixed time slice $t$ to $t+T$, and $\mathcal{M}_{t:t+T}$ sharing the same size with $\mathcal{X}_{t:t+T}$ represents the missing position of the observed data as follows:
\begin{align}
    \mathcal{M}_{\tau,i}=
    \begin{cases}
    1,~\text{if } \mathcal{N}_i \text{ at time slice } \tau  \text{ is observed,}\\
    0,~\text{if } \mathcal{N}_i \text{ at time slice } \tau  \text{ is not observed.}
    \end{cases}
\end{align}
Given $\mathcal{X}_{t:t+T}$, $\mathcal{M}_{t:t+T}$ and the topology graph $\mathcal{G}^K$ which represents a K-nearest neighbours graph structure of traffic networks, the purpose of our paper is to learn a function $Imputer$ to infer the missing values in $\mathcal{X}_{t:t+T}$, that is, to obtain the imputed sequence data $\mathcal{\hat{Y}}_{t:t+T} \in \mathbb{R}^{T \times N \times C}$.
The task can be formulated as:
\begin{align}
    \mathcal{\hat{Y}}_{t:t+T}=Imputer(\langle \mathcal{X}_{t:t+T},\mathcal{M}_{t:t+T},\mathcal{G}^K \rangle|\boldsymbol{\Theta}),
\end{align}
where $\boldsymbol{\Theta}$ denotes the learnable parameters in our model.

\section{Methodology}
\label{sec_Meth}
\begin{figure*}[t]
\centering
\includegraphics[width=\textwidth]{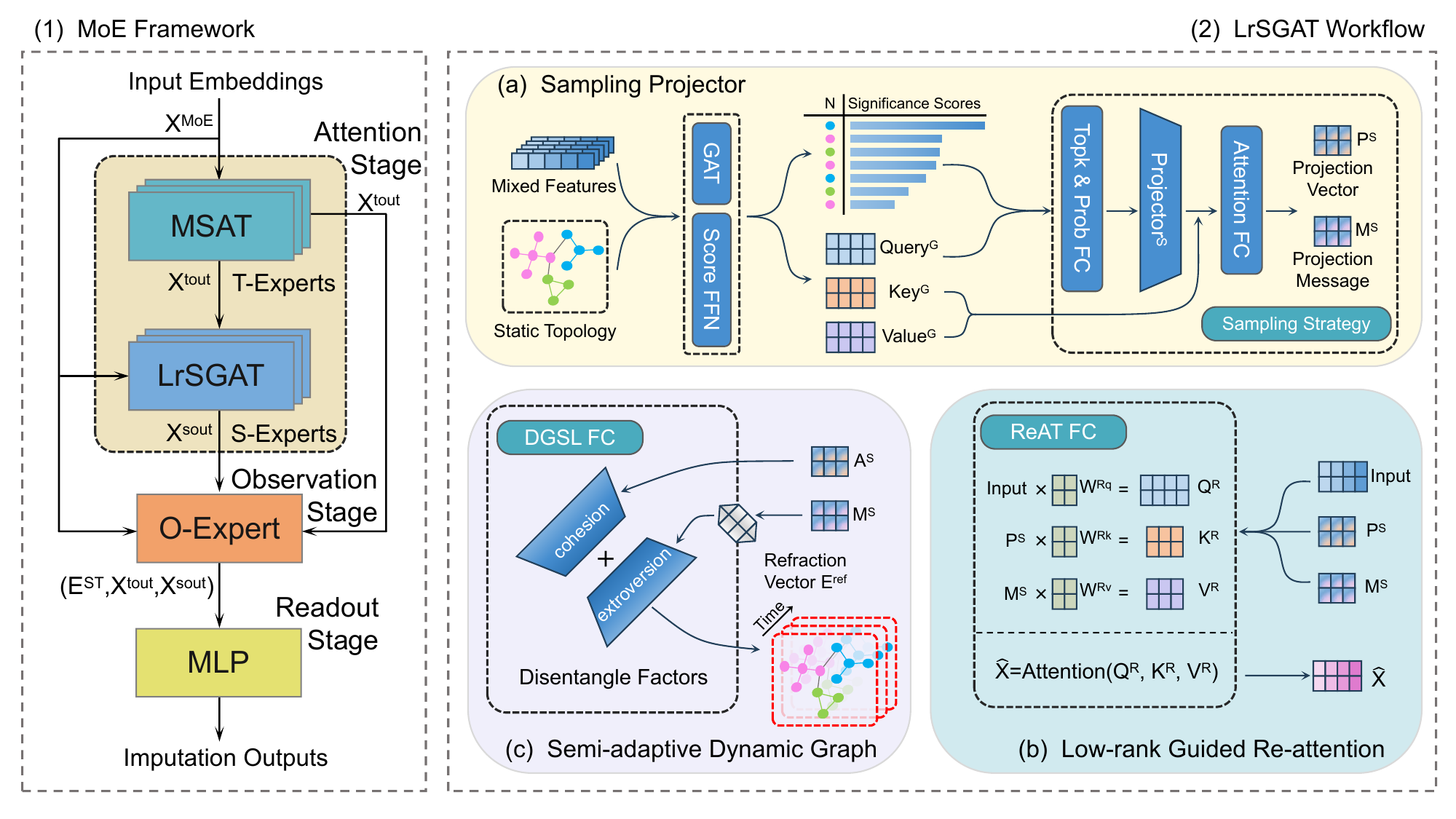}
\caption{The overview of STAMImputer and LrSGAT: 
(1) The framework of MoE contains multi-head self-attention networks as Temporal Experts (T-Experts), LrSGAT networks as Spatial Experts (S-Experts), a feed-forward network as Observation Expert (O-Expert) and a Readout MLP.
(2) The workflow of LrSGAT:
(a) sampling projector generates projection attention with diverse sampling strategies;
(b) low-rank guided re-attention mechanism with projection attention;
(c) dynamic graphs generation with projection attention.
}
\label{Fig_Framework}
\end{figure*}
In this section, we introduce our model STAMImputer.
The overview of STAMImputer is shown in Figure.\ref{Fig_Framework}
The left side of the figure shows the framework of MoE, which depicts the collaboration mode between the observation and the attention experts.
The right side of the figure shows the workflow of the spatial experts, which we introduce in detail in Section~\ref{sec_spa}.

\subsection{Observation Expert and MoE Framework}
As an emerging architecture, MoE has been widely used and discussed in deep learning fields.
Unlike the well-known MoE in~\cite{fedus2022switch,zhang2021moefication}, we extend the expert networks to the outer layer of the framework.
The advantage of this design is its ability to decouple data feature in a controlled way, enabling the temporal and spatial experts to effectively capture feature propagation relationships in both dimensions.
It is called controlled decoupling because there exist a macro-control expert (i.e., Observation Expert in MoE) who can evaluate the trust of the learning results of functional experts by focusing on raw data and sparse features.
Through the tasks and cooperation of experts, the system can decouple dimensional attention while ensuring cohesion in the final result, which is the original intention of the design of our MoE framework.
In summary, the use of MoE enhances the model’s stability and generalization. Regardless of the sparsity in time or space dimensions, the expert network can effectively guide the model based on real-time data characteristics before producing the output.

\paragraph{Observation Expert}
The evaluation ability of observation expert is directly related to the quality of the learning results and the robustness of the model.
The observation expert can learn downstream-oriented evaluation feedback and make trust decisions on the spatio-temporal attention output in the completion task based on the real data (including missing) and sparsity features.
Compared with the structural design of the attention expert, the observation expert can not only achieve the effect of a residual network but also improve the dynamics of residual connections, thereby improving the stability when handling sparse patterns of real complex data.
In addition, it is worth noting that different dimensions of attention have different strengths and weaknesses under different missing patterns.
For example, in the case of random data missing, temporal attention performs better than spatial attention in completing the task, while the opposite is true in the case of block missing patterns ~\cite{marisca2022learning}.
Therefore, observing the expert's dynamic evaluation and sparse sensitivity of inputs improves the model's flexible perception of complex spatio-temporal networks.

\paragraph{MoE Framework} As shown in Figure \ref{Fig_Framework}.1, MoE framework is structured into three main stages:
\textbf{1)} attention stage: to extract spatio-temporal correlations with MSAT and LrSGAT networks;
\textbf{2)} observation stage: to score the credibility of attention experts and weight the outputs of the attention stage;
\textbf{3)} readout stage: to produce the final predictions through Multilayer Perceptron (MLP).

Firstly, a vectors group $\mathcal{X}^{MoE}=\langle\mathcal{X}^{oe} \Vert \mathcal{X}^{in}\rangle$ , in which $\mathcal{X}^{oe} \in \mathbb{R}^{T \times N \times D_{oe}}$ is as inputs for observation expert while $\mathcal{X}^{in} \in \mathbb{R}^{T \times N \times D_{in}}$ is as inputs for attention experts (refer to Section.\ref{Meth_Feat} for details).
Then, hierarchical iterative correlation learning and feature reconstruction by spatio-temporal experts in the attention processing stage:
\begin{align}
    &\mathcal{X}^{tout_i}=Atten^T(\mathcal{X}^{tin_i}),\\
    &\mathcal{X}^{sout_i}=Atten^S(\langle\mathcal{X}^{sin_i} \Vert \mathcal{X}^{tout_i}\rangle).
\end{align}
$\mathcal{X}^{i}$ represents the inputs or outputs of the attention stage in layer ${i}$ and $\mathcal{X}^{out_i}$ will be the input of layer ${i+1}$. 
Obviously, $\mathcal{X}^{tin_0}=\mathcal{X}^{sin_0}=\mathcal{X}^{in}$.
The observation expert performs attention confidence assessment on node ${n}$ at time ${t}$:
\begin{align}
    e^{ST}_{n,t}={softmax}(x^{oe}_{n,t}W^{oe}),
\end{align}
where $W^{oe}\in\mathbb{R}^{D^{oe} \times N^{ae}} $ is a  learnable parameter matrix and ${N}^{ae}$ is the number of attention experts.
Finally, the learning feedback from the experts is weighted and concatenated according to the confidence scores $\mathcal{E}^{ST} \in \mathbb{R}^{T \times N \times N^{ae}}$ from the observation expert and input into the readout stage to obtain the imputation outputs.

\subsection{Attention Experts of Dual-dimensions}
Self-attention extraction and influence propagation are the core of STAMImputer.
Whether the potential interactive relationship can be extracted is the key to improving the imputation effects.
In this section, we describe in detail the design of attention experts in STAMImputer and how they effectively extract latent patterns.

\subsubsection{Temporal Attention Expert Networks}
In the temporal dimension, we introduce the encoder layer of the classic self-attention Transformer~\cite{vaswani2017attention} as the temporal expert network unit, that is, the \textbf{M}ulti-head \textbf{S}elf-\textbf{AT}tention (MSAT) network shown in Figure \ref{Fig_Framework}.1.
In long or short time slices, the interaction patterns of time points are relatively stable (relative to the spatial dimension), which is the main reason why simple time series imputation has a weak ability to perceive high-frequency and struggles to detect abnormal events.
However, temporal self-attention still has a strong advantage in restoring stable sequence trends (i.e., low-frequency recovery).
Using the classic self-attention encoder as the initial  temporal processing layer of the MoE framework can quickly pre-reconstruct missing input features, especially for data with low missing rates or simple missing patterns (such as random point missing).
Moreover, its output vectors can provide richer information for the spatial attention expert network, enabling the dynamic discovery of potential spatial associations.

\subsubsection{Spatial Attention Expert Networks}
\label{sec_spa}
The spatial patterns that can be learned will more closely align with the unpredictable nonlinear correlations observed in real-world networks.
In the spatial attention expert network, as shown in Figure \ref{Fig_Framework}.2, we propose a \textbf{L}ow-\textbf{r}ank guided \textbf{S}ampling \textbf{G}raph \textbf{AT}tention mechanism, namely \textbf{LrSGAT}.
This innovative design employs hybrid sampling strategies to effectively balance local and global correlations across different time periods.
Additionally, the sampled attention weights can also serve as projection vectors for the input to the re-attention layer and be used to construct semi-adaptive dynamic graphs offering dynamic spatial dependencies for downstream traffic intelligence tasks.

\paragraph{Sampling Projector}
We state that the design is to optimize the balance between local and global correlations of spatial nodes while maintaining manageable complexity.
We propose a sampling-based projector design, where a reduced-dimensional attention vector is sampled through a general Graph Attention Network (GAT) guided by a static topology.
The output vector is derived by using the sampled attention as the projection message.
Specifically, an original input $\mathcal{X}^{Sin}_t \in \mathbb{R}^{N \times D^{in}}$ enters the LrSGAT network and extracts the attention weights on the local graph through the static topological adjacency matrix as formula:
\begin{align}
    \mathcal{E}^G_t=softmax\left(\frac{Q^G_t(K^G_t)^\top}{\sqrt{d_k}}\right),
\end{align}
where $Q^G_t \in \mathbb{R}^{N \times 1 \times D^{in}}$ and $K^G_t \in \mathbb{R}^{N \times E \times D^{in}}$ are expanded query and key vectors calculated with input $\mathcal{X}^{Sin}_t$ and the static topology with $E$ denoting the number of neighbors for each node.
In addition, the canonical vectors $Q_t,~K_t,~\text{and}~V_t$ are also obtained in this process. 
$\mathcal{E}^G_t \in \mathbb{R}^{N \times E}$ denotes the feedback of local attention, and its role is to reversely inform the global influence of the node at the current time $t$. 
We set up a trainable scorer vector $\mathcal{W}^{sc} \in \mathbb{R}^{E \times 1}$ to generate a significance score matrix $\mathcal{E}^W_t \in \mathbb{R}^{N \times 1}$ from local attention and perform node sampling based on it:
\begin{align}
    \mathcal{E}^W_t=\mathcal{E}^G_t\mathcal{W}^{sc},
    ~\mathcal{N}^S_t=\langle\mathcal{N}^T_t \Vert \mathcal{N}^U_t\rangle.
\end{align}
$\mathcal{N}^S_t$ denotes the set of nodes sampled at time $t$, where $\mathcal{N}^T_t$ represents the $S=\lceil logN \rceil$  nodes with the highest significance scores, and $\mathcal{N}^U_t$ represents the $S$ nodes of the probability distribution sampling from the remaining parts.
The combination of these two sampling strategies is designed to focus attention on different score domains, ensuring that potential traffic hubs are not overlooked due to data missing.

The abstract concept of $\mathcal{N}^S_t$ refers to the selection of nodes with the highest global influence as well as probabilistically selected random nodes with incremental effects.
Finally, the projection vector consists of the mapping of the attention query or key to the samples, and the projection message is calculated using the sampled query vector via the self-attention mechanism:
\begin{align}
    &Q_t \xrightarrow{N_t^S} P_t^S, \quad K_t \xrightarrow{N_t^S} K_t^S,
    \label{for_kst}\\
    &M_t=softmax\left(\frac{P^S_t(K_t)^\top}{\sqrt{d_k}}\right)V_t.
\end{align}
Through sample mapping, the total size of the obtained projection vector is significantly reduced compared to the original input vector since $S \ll N$.
Furthermore, due to the evaluation of the significance scores and hybrid sampling, representative information is preserved in the projection vector, allowing $P^S_t$ and $M_t$ to pass to the next layer of LrSGAT to preserve the potential local and global correlations of the original input vector.

\paragraph{Low-rank Guided Re-attention}
Low-rank factorization is to approximate a vector $\mathcal{X} \in \mathbb{R}^{m \times n}$ by a vector product representation as $\mathcal{X} = \mathcal{UV}^\top$, where $\mathcal{U} \in \mathbb{R}^{m \times k}$ and $\mathcal{V} \in \mathbb{R}^{n \times k}$ are rank matrices $k \ll \min(m, n)$. 
Factorization seeks to simplify information by breaking down complex network structures into abstract, compressed features that effectively capture the overall structure.
At this layer, as shown in Figure~\ref{Fig_Framework}.2(b), we apply \textbf{Re}-\textbf{AT}tention (ReAT) learning by utilizing the projection vector and the projection message received from the previous layer of the network:
\begin{align}
    &\mathcal{\hat{X}}_t^{s}=Atten^P(\mathcal{X}_t,P^S_t,M^S_t).
\end{align}
$P^S_t$ encapsulates the current spatial network information into a concise and compact representation of its state and generates the key vector to guide the global nodes to pay attention to the critical node information sampled.
$M^S_t$ corresponds to the reference for recovering compressed information $P^S_t$ and generates a value vector to recover the attention guided by the compressed information in the global reconstructed hidden state.
The key to approximate low-rank factorization of the original vector is the projector, which realizes information cohesion through a sampling mechanism and generates corresponding projection messages to guide feature reconstruction.
The design of repeated attention compresses and restores the high-order, low-rank spatial matrix and completes the missing data features in the process.

\paragraph{Semi-adaptive Dynamic Graph}
The pattern of spatial sampled attention and the low-rank factorization approach bring to mind a classic method for constructing adaptive adjacency matrices $\tilde{A}^{adp} \in \mathbb{R}^{N \times N}$~\cite{wu2019graph}, which represent fully connected adjacency relationships learned entirely by the neural network. 
The matrices $\tilde{A}^{adp}$ are formulated as:
\begin{align}
    \tilde{A}^{adp}=softmax(ReLU(E_1E_2^\top)),
\end{align}
where $E_1~\text{and}~E_2 \in \mathbb{R}^{N \times C}$ are learnable parameters.
With $C \ll N$, the constructed adaptive adjacency relationship is achieved by abstracting a cohesive representation and an extroversion representation, compressing and reconstructing the global network for correlation mining and coupling it with the projector model in our two upper layers of the network.
However, it is not difficult to see that each of $E_i$ is completely abstract and static, and its compact approximation does not have any actual nodes to map.
And they are susceptible to low-quality and unreliable structure at data missing conditions~\cite{zou2023se}.

In contrast, the sampling matrix constructed by the projector is dynamic and node-supported.
If we similarly set an $E$ to calculate $\tilde{A}^{adp}=A^SE^
\top$, where $A^S \in \mathbb{R}^{N \times S}$ is the sampled attention vector and $E^\top \in \mathbb{R}^{S \times N}$ is the adaptive matrix, the reconstruction task that $E$ needs to perform is faced with compressed information that changes over time. 
Therefore, simply reconstructing the sampled projections is not of practical significance.
Taking the above factors into account, the \textbf{D}ynamic \textbf{G}raph \textbf{S}tructure \textbf{L}earning (DGSL) function is defined as follows:
\begin{align}
    &\tilde{A}_t^{adp}=softmax(ReLU(A_t^SE^{adp})),\\
    &A_t^S=softmax\left(\frac{Q_t(K_t^S)^\top}{\sqrt{d_k}}\right),\\
    &E^{adp}=toph(M_t(E^{ref})^\top),
\end{align}
where $Q_t \in \mathbb{R}^{N \times D^{in}}$ is the query vector from the Sampling Projector layer, $K_t^S \in \mathbb{R}^{S \times D^{in}}$ is the sampled key vector as in Formula (\ref{for_kst}), $M_t \in \mathbb{R}^{S \times D^{in}}$ is the projection message, $E^{ref} \in \mathbb{R}^{N \times D^{in}}$ is a learnable parameter matrix and $toph(\cdot)$ is the function setting terms less than the median to zero.

\subsection{Spatio-temporal Representation Learning}
\label{Meth_Feat}
In this section, we summarize the spatio-temporal representation learning stage in STAMImputer.
Incorporating additional feature dimensions can accelerate the convergence of expert networks.
In general, the spatio-temporal and frequency features of the original input are embedded before entering the MoE networks.


\begin{table*}[h]
    \centering
    \small
    \begin{tabular}{l|c@{\hskip 5.5pt}c|c@{\hskip 5.5pt}c|c@{\hskip 5.5pt}c|c@{\hskip 5.5pt}c|c@{\hskip 5.5pt}c|c@{\hskip 5.5pt}c|c@{\hskip 5.5pt}c|c@{\hskip 5.5pt}c}
    \toprule
     \multirow{3}{*}{Methods}& \multicolumn{8}{c|}{Point Missing (Missing Rate)} & \multicolumn{8}{c}{Block Missing (Failure Probability)}\\
     \cmidrule{2-17}
     & \multicolumn{2}{c|}{PemsD8} & \multicolumn{2}{c|}{SZ-Taxi} & \multicolumn{2}{c|}{DiDi-SZ} & \multicolumn{2}{c|}{NYC-Taxi}& \multicolumn{2}{c|}{PemsD8} & \multicolumn{2}{c|}{SZ-Taxi} & \multicolumn{2}{c|}{DiDi-SZ} & \multicolumn{2}{c}{NYC-Taxi}\\
     \cmidrule{2-17}
     & 25\% & 60\% & 25\% & 60\% & 25\% & 60\% & 25\% & 60\% & 0.2\% & 1\% & 0.2\% & 1\% & 0.2\% & 1\% & 0.2\% & 1\%  \\
     \midrule
     Mean & 89.51 & 87.32 & 7.93 & 7.96 & 7.68 & 7.70 & 41.07 & 40.41 & 88.03 & 88.01 & 7.60 & 7.69 & 7.81 & 7.65 & 44.96 & 42.04 \\
     KNN & 84.74 & 85.10 & 6.09 & 6.11 & 8.03 & 8.04 & 18.62 & 18.38 & 88.51 & 85.98 & 6.44 & 5.90 & 7.97 & 8.16 & 19.49 & 19.58 \\
     LATC\tablefootnote{The LATC method is invalid for training on SZ-Taxi, so its result is empty.} & 17.46 & 17.53 & - & - & 2.16 & 2.22 & 5.56 & 7.14 & 28.99 & 43.21 & - & - & 2.96 & 5.70 & 7.93 & 8.08 \\
     VAR & 17.48 & 18.82 & 3.57 & 3.42 & 2.50 & 2.53 & 9.86 & 11.97 & 25.29 & 23.82 & 3.87 & 3.72 & 3.04 & 2.62 & 14.05 & 10.97 \\
     rGAIN & 16.33 & 20.81 & 3.49 & 3.65 & 2.16 & 2.20 & 8.26 & 10.16 & 22.92 & 27.29 & 3.79 & 3.75 & 2.29 & 2.30 & 8.96 & 11.61 \\
     BRITS & 15.78 & 18.60 & 3.32 & 3.34 & 2.15 & 2.20 & 6.84 & 7.49 & 16.37 & 21.06 & 3.43 & 3.46 & 2.27 & 2.27 & 7.01 & 7.43 \\
     Transformer & 12.58 & 13.75 & 3.20 & 3.29 & 1.63 & 1.81 & 5.46 & 6.40 & 33.51 & 53.91 & 4.19 & 4.99 & 3.08 & 4.07 & 14.49 & 17.37 \\
     SAITS & 15.12 & 21.66 & 3.30 & 3.35 & 2.13 & 2.05 & 6.27 & 7.28 & 21.48 & 24.79 & 3.41 & 3.52 & 2.22 & 2.18 & 7.18 & 7.49 \\
     SPIN & 15.02 & 13.94 & 3.56 & 3.22 & \underline{1.61} & \underline{1.77} & 6.25 & 5.68 & 15.49 & 20.00 & 3.56 & 3.50 & 2.05 & 2.21 & 7.70 & 6.25 \\
     ImputeFormer & \textbf{11.01} & \underline{13.09} & \underline{3.19} & \underline{3.20} & 1.68 & 2.00 & \underline{5.29} & \underline{5.56} & \textbf{15.07} & \textbf{17.56} & \underline{3.40} & \underline{3.30} & \underline{1.98} & \underline{2.05} & \underline{6.37} & \underline{6.14} \\
     \midrule
     STAMImputer & \underline{11.41} & \textbf{12.32} & \textbf{3.12} & \textbf{3.11} & \textbf{1.57} & \textbf{1.73} & \textbf{4.85} & \textbf{5.34} & \underline{15.38} & \underline{18.06} & \textbf{3.32} & \textbf{3.28} & \textbf{1.94} & \textbf{2.03} & \textbf{5.66} & \textbf{5.81} \\
    \bottomrule 
    \end{tabular}
    \caption{Results (in terms of MAE) on PemsD8, SZ-Taxi, DiDi-SZ, NYC-Taxi benchmarks. 
    The block missing with 0.2\% and 1\% failure probabilities correspond to approximate total missing rates of 10\% and 30\%.}
    \label{tab2}
\end{table*}

\paragraph{Discrete Wavelet Transform}

Discrete wavelet transform (DWT) and inverse wavelet transform (IWT) are powerful mathematical tools that decompose signals into different frequency components, effectively capturing sequence characteristics.
Specifically, given the original observed sequence vector $\mathcal{X}_{t:t+T} \in \mathbb{R}^{N \times T}$, and selecting the wavelet basis $w$ and decomposition level $j$, DWT decomposes the sequence $\mathcal{X}^{in}$ into low-frequency and high-frequency components:
\begin{align}
    DWT(\mathcal{X}^{in},~w,~j)=\{C_0,~C_1,~\ldots,~C_j\},
\end{align}
where $C_0$ denotes low-frequency components, and the rest denote high-frequency components.
IWT is formulated as:
\begin{align}
    &\mathcal{X}^l=IWT(\{C_0,~0,~\ldots,~0\},~w,~j),\\
    &\mathcal{X}^h=IWT(\{0,~C_1,~\ldots,~C_j\},~w,~j),
\end{align}
where $\mathcal{X}^l \in \mathbb{R}^{N \times T}$ and $\mathcal{X}^h \in \mathbb{R}^{N \times T}$ share the same shape with original vector $\mathcal{X}^{in}$.

\paragraph{Spatio-temporal Embedding}
Firstly, the temporal feature, as one of the fundamental characteristics, encompasses both the data collection time and the associated weekday information, denoted as $P^u \in \mathbb{R}^{N \times T \times 2}$.
Secondly, the spatio-temporal position encoding employs a learnable parameter matrix $P^{st} \in \mathbb{R}^{T \times N \times D^{pe}}$, where the depth of its dimensions offers richer feature references for the data. 
Thirdly, statistical learning is used to calculate the sparsity rate in both time and space dimensions as the sparse features denoted $P^{sp} \in \mathbb{R}^{N \times T \times 2}$, 
which enable the observation expert to evaluate the attention confidence scores.

Finally, We adopt a dimension expansion $\mathrm{MLP}$ layer to initialize spatio-temporal and frequency features:
\begin{align}
    &\mathcal{X}^{MoE}=\langle\mathcal{X}^{oe} \Vert \mathcal{X}^{in}\rangle,~\mathcal{X}^{oe}=\langle\mathcal{X}_{t:t+T } \Vert P^{sp}\rangle,\\
    &\mathcal{X}^{in}=\langle\mathrm{MLP}(\mathcal{X}_{t:t+T},~\mathcal{X}^l,~\mathcal{X}^h,~P^u) \Vert P^{st}\rangle,
\end{align}
where $\mathcal{X}^{MoE}$ is the initialized vector that will act as embedded inflow of our STAMImputer models.
\section{Experiment}
\label{sec_exp}

\paragraph{Experimental Setup}
The detailed spatio-temporal feature information of the selected benchmark datasets is shown in Table~\ref{tab1}.
For missing patterns, we consider two scenarios discussed in previous work~\cite{cini2021filling}, point missing and block missing.
Since we aim to perform missing imputations with spatio-temporal attention on sparse traffic data, we picked ImputerFormer as the state-of-the-art baseline method.
Imputerformer introduces a low-rank projection in the time dimension for the first time, then adopts a learnable embedding representation in the space dimension.
We then consider other representative deep learning methods:
1) SPIN~\cite{marisca2022learning}: an attention-based model with sparse spatio-temporal graphs.
2) SAITS~\cite{du2023saits}: a diagonally masked self-attention Transformer with a weighted combination of representations.
3) Transformer~\cite{vaswani2017attention}: canonical Transformer with self-attention mechanism.
4) BRITS~\cite{cao2018brits}: An imputation method that directly learns missing values in a bi-recurrent dynamical system without any specific assumptions.
5) rGAIN~\cite{yoon2018gain}: A temporal imputation model based on a generative adversarial net architecture.
Finally, we consider classic statistical and optimization models for comparison:
1) Mean: imputation of observation average values.
2) KNN: ${K}$ neighbor mean value imputation.
3) LATC~\cite{chen2021low}: completion of the low-rank autoregressive tensor.
4) VAR: a vector autoregressive one-step-ahead predictor.
All experiments are conducted on a server with an Intel(R) Xeon(R) Platinum 8336C CPU operating at 2.30GHz and an NVIDIA A800 GPU with 80GB of memory for the above models and datasets.

\begin{table}[h]
    \centering
    \small
    \begin{tabular}{l@{\hskip 16.5pt}l@{\hskip 16.5pt}l@{\hskip 16.5pt}l@{\hskip 16.5pt}l}
        \toprule
        Benchmark  & Steps & Nodes & Interval & Type \\
        \midrule
        PemsD8 & 17856 & 170 & 5min & Traffic flow \\
        SZ-Taxi & 2976 & 156 & 15min & Traffic speed \\
        DiDi-SZ & 17280 & 627 & 10min & Traffic speed \\
        NYC-Taxi & 1848 & 263 & 60min & Traffic flow \\
        \bottomrule
    \end{tabular}
    \caption{Benchmark datasets details.}
    \label{tab1}
\end{table}

\subsection{Performance Results Comparison}
\subsubsection{Imputation Task}
The imputation results on the four benchmarks are shown in Table~\ref{tab2}.
For the point-missing pattern, we choose two random missing rates of 25\% and 60\%, and for the block-missing pattern, we choose two failure probabilities of 0.2\% and 1\% for the $S \sim U(12,~48)$ steps, both with a random missing rate of 5\%.
STAMImputer generally achieves the best performance, except in a few cases where it ranks second in all experiments. 
Especially in urban travel benchmarks our method demonstrates more significant advantages.
ImputeFormer and SAITS show a strong perception of temporal attention. 
Although ImputeFormer has a corresponding attention module in the spatial dimension, both methods are weak for spatially more complex urban traffic or datasets with longer temporal steps.
SPIN focuses on spatial sparse graph attention and has also shown competitiveness in complex urban traffic speed data imputation.
In the case of data from the urban traffic network with many nodes, such as DiDi-SZ, the imputation error of SPIN is even better than that of ImputeFormer. 
Still, its spatio-temporal complexity is higher when faced with complex road networks.
Transformer with a classic self-attention model is ideal for completing point-missing patterns but is relatively ineffective for blocking missing patterns.
Compared with the above deep learning models, the statistical and optimization methods are less competitive in imputation effect due to their limited capabilities.

\begin{table}[h]
    \centering
    \small
    \begin{tabular}{l|c@{\hskip 6pt}c@{\hskip 6pt}c|c}
        \toprule
        \multirow{2}{*}[-1ex]{Scenarios} & \multicolumn{3}{c|}{Point Missing} & \multirow{2}{*}[-1ex]{\makecell[c]{Block\\Missing}} \\
        \cmidrule{2-4}
        & 25\% & 40\% & 60\% &\\
        \midrule
        GWNet+Full &\multicolumn{4}{c}{7.75}\\
        \midrule
        GWNet+Sparse & 8.40 & 9.15 & 9.54 & 10.60 \\
        GWNet+STAMImputer & 8.39 & 8.38 & 8.56 & 8.77 \\
        \midrule
        GWNet+STAMImputer+DGSL & \textbf{8.03} & \textbf{8.02} & \textbf{8.12} & \textbf{8.25} \\
        \bottomrule
    \end{tabular}
    \caption{Prediction results of Graph-Wavenent model(in terms of MAE) on NYC-Taxi benchmark.}
    \label{tab3}
\end{table}

\subsubsection{Downstream Task with Dynamic Graph}
To verify the effectiveness of the DGSL layer results, we conduct additional experiments with a classic Graph-Wavenet network~\cite{wu2019graph} for downstream prediction tasks.
Table~\ref{tab3} shows the effects of the downstream prediction task in multiple experimental scenarios on the NYC-Taxi benchmark.
To verify the information gain of dynamic graphs compared to static topologies, we disabled the fully adaptive adjacency matrix in the Graph-Wavenet network.
We perform traffic prediction experiments under different random point missing rates and block missing scenarios by comparing the unmasked model learning results.
Graph-Wavenet can cope to a certain extent with its excellent spatio-temporal perception in lower random missing rates. 
Still, when the random missing rate increases or blocks are missing, the prediction performance is greatly disturbed.
We test the improvement effect of STAMImputer on the model in two steps:
1) adding a pre-trained imputation model; 
2) adding a pre-trained imputation model with its learnable dynamic graph module.
As shown in Table \ref{tab3}, our imputation model enhances the downstream prediction in failed sparse scenarios (i.e. high sparsity or missing block).

\subsection{Robustness Analysis}
\begin{figure}[t]
\centering
\includegraphics[width=\columnwidth]{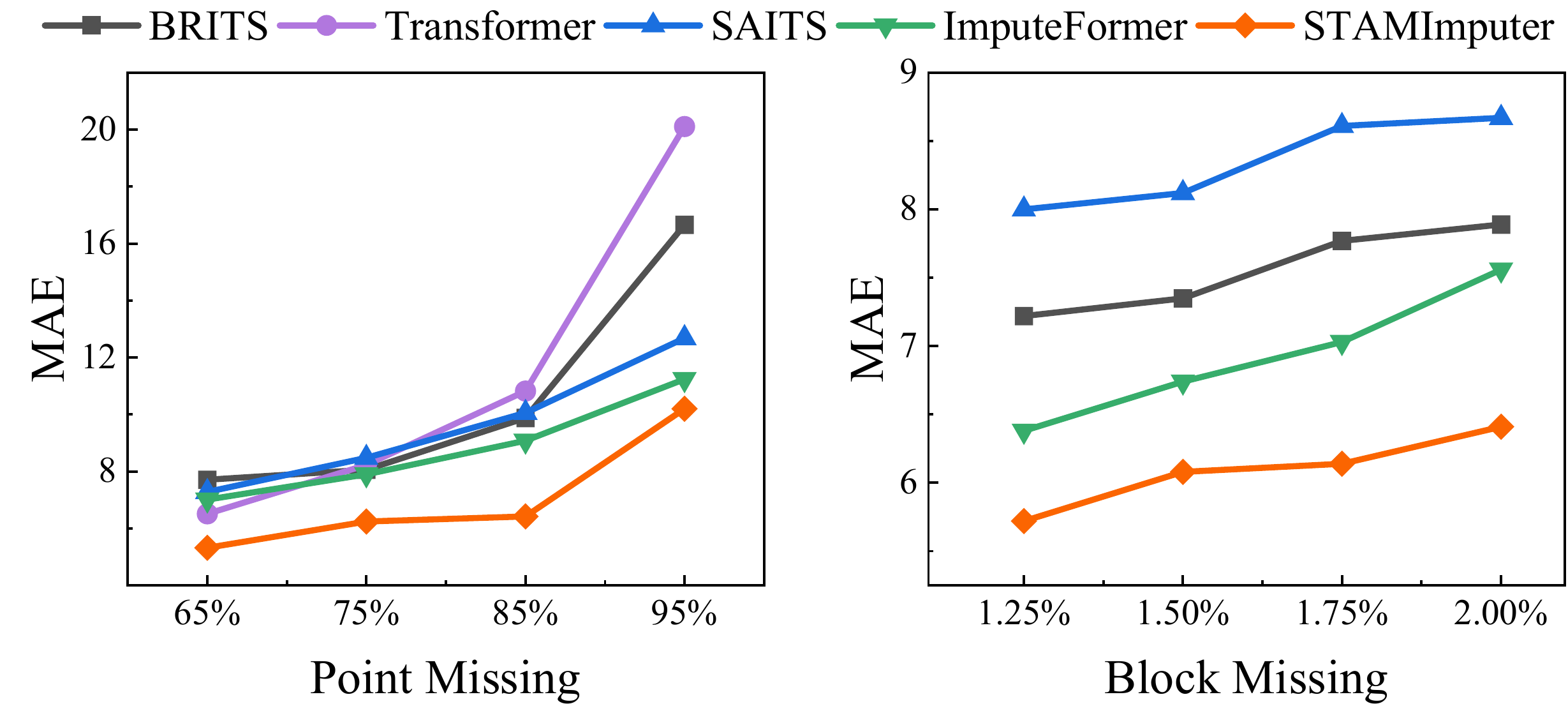}
\caption{
Robustness analysis on NYC-Taxi benchmark.
}
\label{Fig_rob}
\end{figure}

To evaluate the robustness of the model on highly sparse data, we add imputation experiments with varying degrees of sparsity.
The results are shown in Figure~\ref{Fig_rob}.
Since the Transformer is ineffective in blocking missing patterns, its performance curve is not presented.
In general, deep learning completion models suffer performance degradation as sparsity increases.
However, it can be seen that the combination of attention matrices and low-rank constraints in the STAMImputer and ImputeFormer models makes them more robust in the face of high sparsity.
In comparison, STAMImputer exhibits lower imputation errors.

\begin{figure}[h]
\centering
\includegraphics[width=\columnwidth]{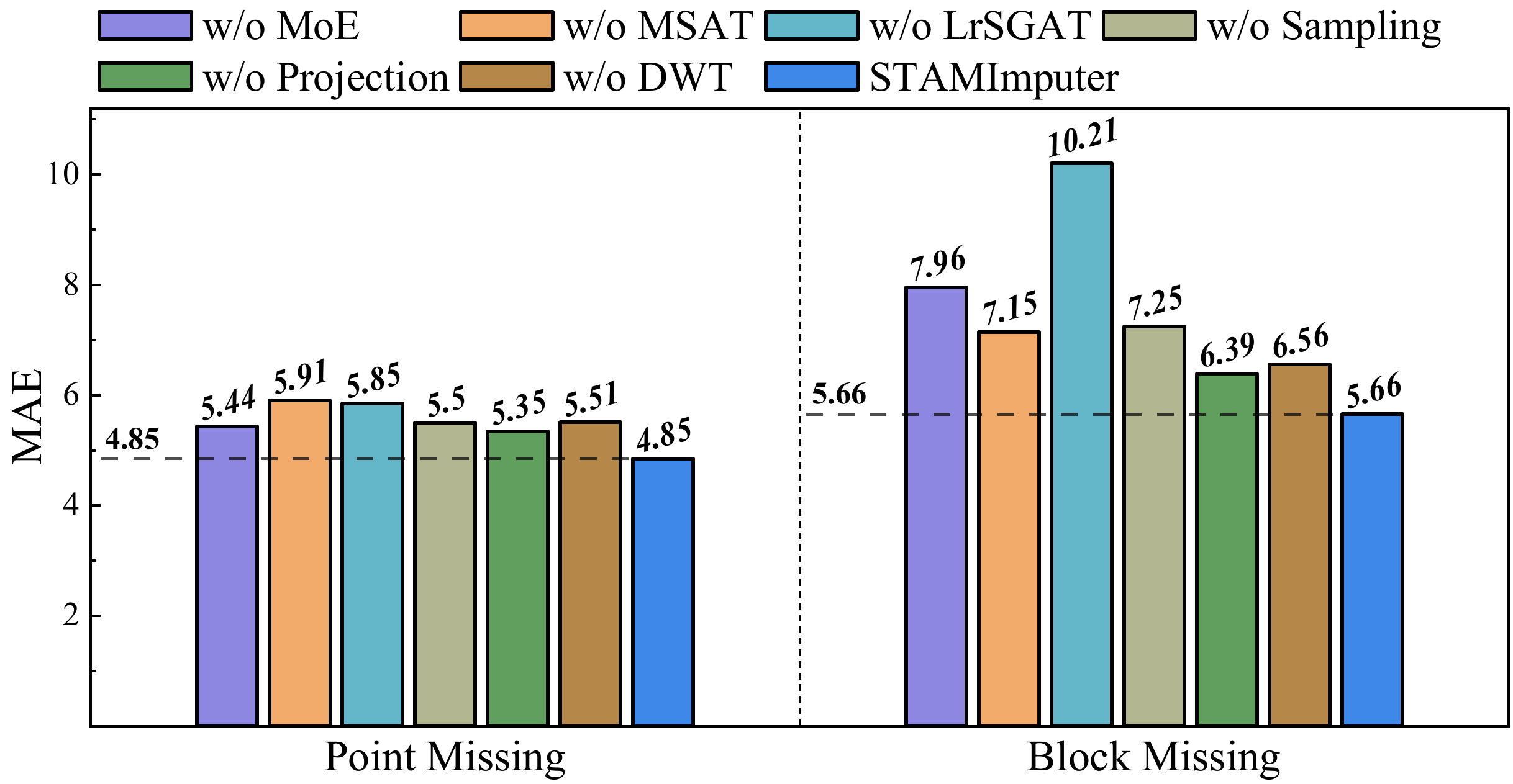}
\caption{
Ablation studies result on NYC-Taxi benchmark.
}
\label{Fig_ablation}
\end{figure}

\subsection{Ablation Study}

We also perform ablation studies to verify the significance of the STAMImputer framework designs.
The imputation results of the ablation experiments are shown in Figure~\ref{Fig_ablation}. 
We first remove the external framework of MoE, in which case we do not consider the macro-control of spatio-temporal attention by observing experts.
When we implement time-to-space sequential framework instead of MoE's regulation of spatio-temporal attention in the ablation study, the model's ability to perceive spatio-temporal associations decreases. 
MoE plays a consistent role in the imputation task in both missing patterns.
Secondly, we compare attention expert networks with four variants: MSAT with MLP as a spatial expert, LrSGAT with MLP as a temporal expert, MSAT with $K$ neighbours GAT as a spatial expert and MSAT with full GAT as a spatial expert.
As shown in Figure~\ref{Fig_ablation}, attention expert networks play the most critical role in STAMImputer, especially in the block missing pattern, where the performance degenerates substantially after replacing LrSGAT with MLP.
We finally remove the wavelet variable from the initial input to verify that the additional dimension added to the embedding helps to extract valuable information for STAMImputer.

\section{Conclusion}
\label{Conclusion}

This paper presents STAMImputer, a traffic data imputation model that leverages MoE framework and dynamically balances spatio-temporal attention. 
We introduce a novel spatial attention sampling mechanism  that combines graph attention with low-rank decomposition to capture spatial correlations. 
Experiments across multiple benchmarks demonstrate that STAMImputer outperforms existing models, and also improves performance in downstream tasks. 
Future research will focus on developing more efficient methods for constructing dynamic graphs.

\section*{Acknowledgments}
\label{Ack}

This work has been supported in part by NSFC through grants 62441612 and 62322202, Local Science and Technology Development Fund of Hebei Province Guided by the Central Government of China through grant 246Z0102G, the ``Pionee'' and ``Leading Goose'' R\&D Program of Zhejiang through grant 2025C02044, National Key Laboratory under grant 241-HF-D07-01, Hebei Natural Science Foundation through grant F2024210008, and CCF-DiDi GAIA collaborative Research Funds for Young Scholars through grant 202422.


\begin{thebibliography}{}

\bibitem[\protect\citeauthoryear{Benkraouda \bgroup \em et al.\egroup }{2020}]{benkraouda2020traffic}
Ouafa Benkraouda, Bilal~Thonnam Thodi, Hwasoo Yeo, Monica Menendez, and Saif~Eddin Jabari.
\newblock Traffic data imputation using deep convolutional neural networks.
\newblock {\em IEEE Access}, 8:104740--104752, 2020.

\bibitem[\protect\citeauthoryear{Cao \bgroup \em et al.\egroup }{2018}]{cao2018brits}
Wei Cao, Dong Wang, Jian Li, Hao Zhou, Lei Li, and Yitan Li.
\newblock Brits: Bidirectional recurrent imputation for time series.
\newblock {\em Advances in neural information processing systems}, 31, 2018.

\bibitem[\protect\citeauthoryear{Chan \bgroup \em et al.\egroup }{2023}]{chan2023missing}
Robin Kuok~Cheong Chan, Joanne Mun-Yee Lim, and Rajendran Parthiban.
\newblock Missing traffic data imputation for artificial intelligence in intelligent transportation systems: review of methods, limitations, and challenges.
\newblock {\em IEEE Access}, 11:34080--34093, 2023.

\bibitem[\protect\citeauthoryear{Chen and Chen}{2022}]{chen2022novel}
Yong Chen and Xiqun~Michael Chen.
\newblock A novel reinforced dynamic graph convolutional network model with data imputation for network-wide traffic flow prediction.
\newblock {\em Transportation Research Part C: Emerging Technologies}, 143:103820, 2022.

\bibitem[\protect\citeauthoryear{Chen and Sun}{2021}]{chen2021bayesian}
Xinyu Chen and Lijun Sun.
\newblock Bayesian temporal factorization for multidimensional time series prediction.
\newblock {\em IEEE Transactions on Pattern Analysis and Machine Intelligence}, 44(9):4659--4673, 2021.

\bibitem[\protect\citeauthoryear{Chen \bgroup \em et al.\egroup }{2019}]{chen2019traffic}
Yuanyuan Chen, Yisheng Lv, and Fei-Yue Wang.
\newblock Traffic flow imputation using parallel data and generative adversarial networks.
\newblock {\em IEEE Transactions on Intelligent Transportation Systems}, 21(4):1624--1630, 2019.

\bibitem[\protect\citeauthoryear{Chen \bgroup \em et al.\egroup }{2020}]{chen2020nonconvex}
Xinyu Chen, Jinming Yang, and Lijun Sun.
\newblock A nonconvex low-rank tensor completion model for spatiotemporal traffic data imputation.
\newblock {\em Transportation Research Part C: Emerging Technologies}, 117:102673, 2020.

\bibitem[\protect\citeauthoryear{Chen \bgroup \em et al.\egroup }{2021}]{chen2021low}
Xinyu Chen, Mengying Lei, Nicolas Saunier, and Lijun Sun.
\newblock Low-rank autoregressive tensor completion for spatiotemporal traffic data imputation.
\newblock {\em IEEE Transactions on Intelligent Transportation Systems}, 23(8):12301--12310, 2021.

\bibitem[\protect\citeauthoryear{Cichocki and Phan}{2009}]{cichocki2009fast}
Andrzej Cichocki and Anh-Huy Phan.
\newblock Fast local algorithms for large scale nonnegative matrix and tensor factorizations.
\newblock {\em IEICE transactions on fundamentals of electronics, communications and computer sciences}, 92(3):708--721, 2009.

\bibitem[\protect\citeauthoryear{Cini \bgroup \em et al.\egroup }{2021}]{cini2021filling}
Andrea Cini, Ivan Marisca, and Cesare Alippi.
\newblock Filling the g\_ap\_s: Multivariate time series imputation by graph neural networks.
\newblock {\em arXiv preprint arXiv:2108.00298}, 2021.

\bibitem[\protect\citeauthoryear{Deng \bgroup \em et al.\egroup }{2021}]{deng2021graph}
Lei Deng, Xiao-Yang Liu, Haifeng Zheng, Xinxin Feng, and Youjia Chen.
\newblock Graph spectral regularized tensor completion for traffic data imputation.
\newblock {\em IEEE Transactions on Intelligent Transportation Systems}, 23(8):10996--11010, 2021.

\bibitem[\protect\citeauthoryear{Du \bgroup \em et al.\egroup }{2023}]{du2023saits}
Wenjie Du, David C{\^o}t{\'e}, and Yan Liu.
\newblock Saits: Self-attention-based imputation for time series.
\newblock {\em Expert Systems with Applications}, 219:119619, 2023.

\bibitem[\protect\citeauthoryear{Fang \bgroup \em et al.\egroup }{2023}]{fang2023spatio}
Yuchen Fang, Yanjun Qin, Haiyong Luo, Fang Zhao, Bingbing Xu, Liang Zeng, and Chenxing Wang.
\newblock When spatio-temporal meet wavelets: Disentangled traffic forecasting via efficient spectral graph attention networks.
\newblock In {\em 2023 IEEE 39th International Conference on Data Engineering (ICDE)}, pages 517--529. IEEE, 2023.

\bibitem[\protect\citeauthoryear{Fedus \bgroup \em et al.\egroup }{2022}]{fedus2022switch}
William Fedus, Barret Zoph, and Noam Shazeer.
\newblock Switch transformers: Scaling to trillion parameter models with simple and efficient sparsity.
\newblock {\em Journal of Machine Learning Research}, 23(120):1--39, 2022.

\bibitem[\protect\citeauthoryear{Kong \bgroup \em et al.\egroup }{2023}]{kong2023dynamic}
Xiangjie Kong, Wenfeng Zhou, Guojiang Shen, Wenyi Zhang, Nali Liu, and Yao Yang.
\newblock Dynamic graph convolutional recurrent imputation network for spatiotemporal traffic missing data.
\newblock {\em Knowledge-Based Systems}, 261:110188, 2023.

\bibitem[\protect\citeauthoryear{Li \bgroup \em et al.\egroup }{2013}]{li2013efficient}
Li~Li, Yuebiao Li, and Zhiheng Li.
\newblock Efficient missing data imputing for traffic flow by considering temporal and spatial dependence.
\newblock {\em Transportation research part C: emerging technologies}, 34:108--120, 2013.

\bibitem[\protect\citeauthoryear{Li \bgroup \em et al.\egroup }{2022}]{li2022fine}
Jiyue Li, Senzhang Wang, Jiaqiang Zhang, Hao Miao, Junbo Zhang, and S~Yu Philip.
\newblock Fine-grained urban flow inference with incomplete data.
\newblock {\em IEEE Transactions on Knowledge and Data Engineering}, 35(6):5851--5864, 2022.

\bibitem[\protect\citeauthoryear{Liang \bgroup \em et al.\egroup }{2022}]{liang2022memory}
Yuebing Liang, Zhan Zhao, and Lijun Sun.
\newblock Memory-augmented dynamic graph convolution networks for traffic data imputation with diverse missing patterns.
\newblock {\em Transportation Research Part C: Emerging Technologies}, 143:103826, 2022.

\bibitem[\protect\citeauthoryear{Liu \bgroup \em et al.\egroup }{2023}]{liu2023cross}
Zhanyu Liu, Guanjie Zheng, and Yanwei Yu.
\newblock Cross-city few-shot traffic forecasting via traffic pattern bank.
\newblock In {\em Proceedings of the 32nd ACM International Conference on Information and Knowledge Management}, pages 1451--1460, 2023.

\bibitem[\protect\citeauthoryear{Ma \bgroup \em et al.\egroup }{2019}]{ma2019cdsa}
Jiawei Ma, Zheng Shou, Alireza Zareian, Hassan Mansour, Anthony Vetro, and Shih-Fu Chang.
\newblock Cdsa: cross-dimensional self-attention for multivariate, geo-tagged time series imputation.
\newblock {\em arXiv preprint arXiv:1905.09904}, 2019.

\bibitem[\protect\citeauthoryear{Marisca \bgroup \em et al.\egroup }{2022}]{marisca2022learning}
Ivan Marisca, Andrea Cini, and Cesare Alippi.
\newblock Learning to reconstruct missing data from spatiotemporal graphs with sparse observations.
\newblock {\em Advances in Neural Information Processing Systems}, 35:32069--32082, 2022.

\bibitem[\protect\citeauthoryear{Nelwamondo \bgroup \em et al.\egroup }{2007}]{nelwamondo2007missing}
Fulufhelo~V Nelwamondo, Shakir Mohamed, and Tshilidzi Marwala.
\newblock Missing data: A comparison of neural network and expectation maximization techniques.
\newblock {\em Current Science}, pages 1514--1521, 2007.

\bibitem[\protect\citeauthoryear{Nie \bgroup \em et al.\egroup }{2024}]{nie2024imputeformer}
Tong Nie, Guoyang Qin, Wei Ma, Yuewen Mei, and Jian Sun.
\newblock Imputeformer: Low rankness-induced transformers for generalizable spatiotemporal imputation.
\newblock In {\em Proceedings of the 30th ACM SIGKDD Conference on Knowledge Discovery and Data Mining}, pages 2260--2271, 2024.

\bibitem[\protect\citeauthoryear{Van~Buuren and Groothuis-Oudshoorn}{2011}]{van2011mice}
Stef Van~Buuren and Karin Groothuis-Oudshoorn.
\newblock mice: Multivariate imputation by chained equations in r.
\newblock {\em Journal of statistical software}, 45:1--67, 2011.

\bibitem[\protect\citeauthoryear{Vaswani}{2017}]{vaswani2017attention}
A~Vaswani.
\newblock Attention is all you need.
\newblock {\em Advances in Neural Information Processing Systems}, 2017.

\bibitem[\protect\citeauthoryear{Wang \bgroup \em et al.\egroup }{2022}]{wang2022generative}
Senzhang Wang, Jiyue Li, Hao Miao, Junbo Zhang, Junxing Zhu, and Jianxin Wang.
\newblock Generative-free urban flow imputation.
\newblock In {\em Proceedings of the 31st ACM International Conference on Information \& Knowledge Management}, pages 2028--2037, 2022.

\bibitem[\protect\citeauthoryear{Wei \bgroup \em et al.\egroup }{2024}]{wei2024self}
Xiulan Wei, Yong Zhang, Shaofan Wang, Xia Zhao, Yongli Hu, and Baocai Yin.
\newblock Self-attention graph convolution imputation network for spatio-temporal traffic data.
\newblock {\em IEEE Transactions on Intelligent Transportation Systems}, 2024.

\bibitem[\protect\citeauthoryear{Wu \bgroup \em et al.\egroup }{2019}]{wu2019graph}
Zonghan Wu, Shirui Pan, Guodong Long, Jing Jiang, and Chengqi Zhang.
\newblock Graph wavenet for deep spatial-temporal graph modeling.
\newblock {\em arXiv preprint arXiv:1906.00121}, 2019.

\bibitem[\protect\citeauthoryear{Xu \bgroup \em et al.\egroup }{2022}]{xu2022traffic}
Qianxiong Xu, Sijie Ruan, Cheng Long, Liang Yu, and Chen Zhang.
\newblock Traffic speed imputation with spatio-temporal attentions and cycle-perceptual training.
\newblock In {\em Proceedings of the 31st ACM International Conference on Information \& Knowledge Management}, pages 2280--2289, 2022.

\bibitem[\protect\citeauthoryear{Xu \bgroup \em et al.\egroup }{2023}]{xu2023hrst}
Xiuqin Xu, Mingwei Lin, Xin Luo, and Zeshui Xu.
\newblock Hrst-lr: a hessian regularization spatio-temporal low rank algorithm for traffic data imputation.
\newblock {\em IEEE Transactions on Intelligent Transportation Systems}, 24(10):11001--11017, 2023.

\bibitem[\protect\citeauthoryear{Xu \bgroup \em et al.\egroup }{2024}]{xu2024hierarchical}
Dongwei Xu, Hang Peng, Yufu Tang, and Haifeng Guo.
\newblock Hierarchical spatio-temporal graph convolutional neural networks for traffic data imputation.
\newblock {\em Information Fusion}, 106:102292, 2024.

\bibitem[\protect\citeauthoryear{Ye \bgroup \em et al.\egroup }{2021}]{ye2021spatial}
Yongchao Ye, Shiyao Zhang, and James~JQ Yu.
\newblock Spatial-temporal traffic data imputation via graph attention convolutional network.
\newblock In {\em International Conference on Artificial Neural Networks}, pages 241--252. Springer, 2021.

\bibitem[\protect\citeauthoryear{Yi \bgroup \em et al.\egroup }{2016}]{yi2016st}
Xiuwen Yi, Yu~Zheng, Junbo Zhang, and Tianrui Li.
\newblock St-mvl: Filling missing values in geo-sensory time series data.
\newblock In {\em Proceedings of the 25th international joint conference on artificial intelligence}, 2016.

\bibitem[\protect\citeauthoryear{Yin \bgroup \em et al.\egroup }{2021}]{yin2021deep}
Xueyan Yin, Genze Wu, Jinze Wei, Yanming Shen, Heng Qi, and Baocai Yin.
\newblock Deep learning on traffic prediction: Methods, analysis, and future directions.
\newblock {\em IEEE Transactions on Intelligent Transportation Systems}, 23(6):4927--4943, 2021.

\bibitem[\protect\citeauthoryear{Yoon \bgroup \em et al.\egroup }{2018}]{yoon2018gain}
Jinsung Yoon, James Jordon, and Mihaela Schaar.
\newblock Gain: Missing data imputation using generative adversarial nets.
\newblock In {\em International conference on machine learning}, pages 5689--5698. PMLR, 2018.

\bibitem[\protect\citeauthoryear{Yu \bgroup \em et al.\egroup }{2016}]{yu2016temporal}
Hsiang-Fu Yu, Nikhil Rao, and Inderjit~S Dhillon.
\newblock Temporal regularized matrix factorization for high-dimensional time series prediction.
\newblock {\em Advances in neural information processing systems}, 29, 2016.

\bibitem[\protect\citeauthoryear{Yuan \bgroup \em et al.\egroup }{2022}]{yuan2022stgan}
Ye~Yuan, Yong Zhang, Boyue Wang, Yuan Peng, Yongli Hu, and Baocai Yin.
\newblock Stgan: Spatio-temporal generative adversarial network for traffic data imputation.
\newblock {\em IEEE Transactions on Big Data}, 9(1):200--211, 2022.

\bibitem[\protect\citeauthoryear{Zhang \bgroup \em et al.\egroup }{2021}]{zhang2021moefication}
Zhengyan Zhang, Yankai Lin, Zhiyuan Liu, Peng Li, Maosong Sun, and Jie Zhou.
\newblock Moefication: Transformer feed-forward layers are mixtures of experts.
\newblock {\em arXiv preprint arXiv:2110.01786}, 2021.

\bibitem[\protect\citeauthoryear{Zhang \bgroup \em et al.\egroup }{2022}]{zhang2022self}
Yong Zhang, Xiulan Wei, Xinyu Zhang, Yongli Hu, and Baocai Yin.
\newblock Self-attention graph convolution residual network for traffic data completion.
\newblock {\em IEEE Transactions on Big Data}, 9(2):528--541, 2022.

\bibitem[\protect\citeauthoryear{Zhang \bgroup \em et al.\egroup }{2024a}]{zhang2024score}
Shunyang Zhang, Senzhang Wang, Hao Miao, Hao Chen, Changjun Fan, and Jian Zhang.
\newblock Score-cdm: Score-weighted convolutional diffusion model for multivariate time series imputation.
\newblock {\em arXiv preprint arXiv:2405.13075}, 2024.

\bibitem[\protect\citeauthoryear{Zhang \bgroup \em et al.\egroup }{2024b}]{zhang2024comprehensive}
Yimei Zhang, Xiangjie Kong, Wenfeng Zhou, Jin Liu, Yanjie Fu, and Guojiang Shen.
\newblock A comprehensive survey on traffic missing data imputation.
\newblock {\em IEEE Transactions on Intelligent Transportation Systems}, 2024.

\bibitem[\protect\citeauthoryear{Zhao \bgroup \em et al.\egroup }{2020}]{zhao2020traffic}
Junhui Zhao, Yiwen Nie, Shanjin Ni, and Xiaoke Sun.
\newblock Traffic data imputation and prediction: An efficient realization of deep learning.
\newblock {\em IEEE Access}, 8:46713--46722, 2020.

\bibitem[\protect\citeauthoryear{Zou \bgroup \em et al.\egroup }{2023}]{zou2023se}
Dongcheng Zou, Hao Peng, Xiang Huang, Renyu Yang, Jianxin Li, Jia Wu, Chunyang Liu, and Philip~S Yu.
\newblock Se-gsl: A general and effective graph structure learning framework through structural entropy optimization.
\newblock In {\em Proceedings of the ACM Web Conference 2023}, pages 499--510, 2023.

\bibitem[\protect\citeauthoryear{Zou \bgroup \em et al.\egroup }{2024}]{zou2024multispans}
Dongcheng Zou, Senzhang Wang, Xuefeng Li, Hao Peng, Yuandong Wang, Chunyang Liu, Kehua Sheng, and Bo~Zhang.
\newblock Multispans: a multi-range spatial-temporal transformer network for traffic forecast via structural entropy optimization.
\newblock In {\em Proceedings of the 17th ACM International conference on web search and data mining}, pages 1032--1041, 2024.

\end{thebibliography}
\appendix
\section*{Appendix}

\begin{figure}[!t]
    \centering
    \begin{subfigure}[b]{0.45\textwidth}
        \includegraphics[width=\textwidth]{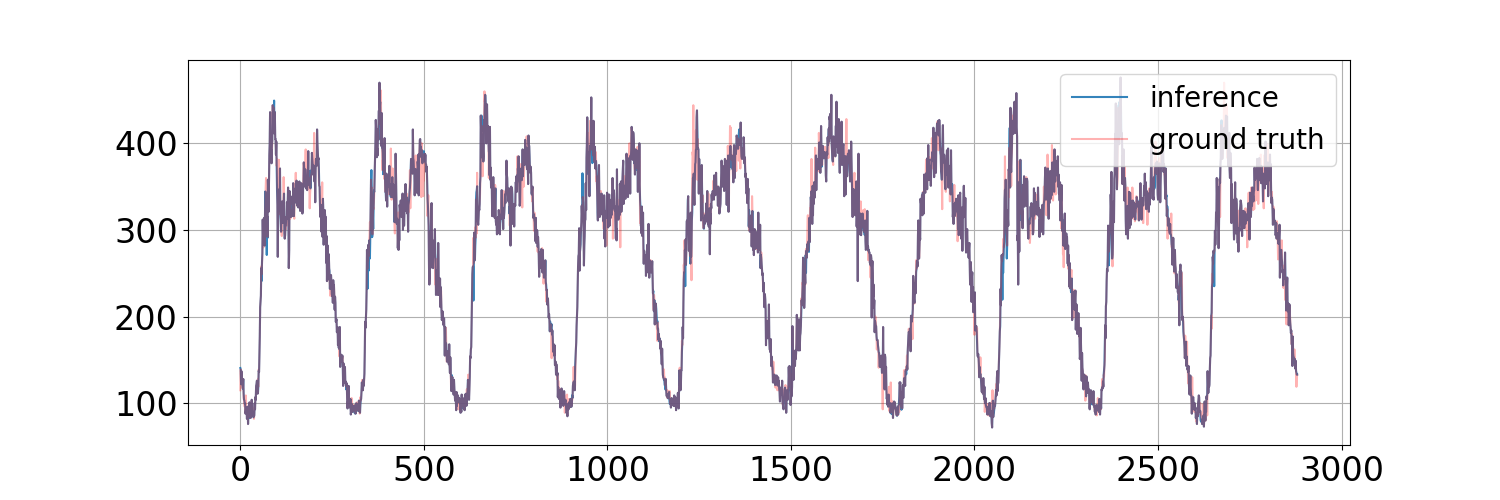}
        \caption{Point missing imputation on PEMSD8 on Node\#2.}
        \label{fig:sub1}
    \end{subfigure}
    \hfill
    \begin{subfigure}[b]{0.45\textwidth}
        \includegraphics[width=\textwidth]{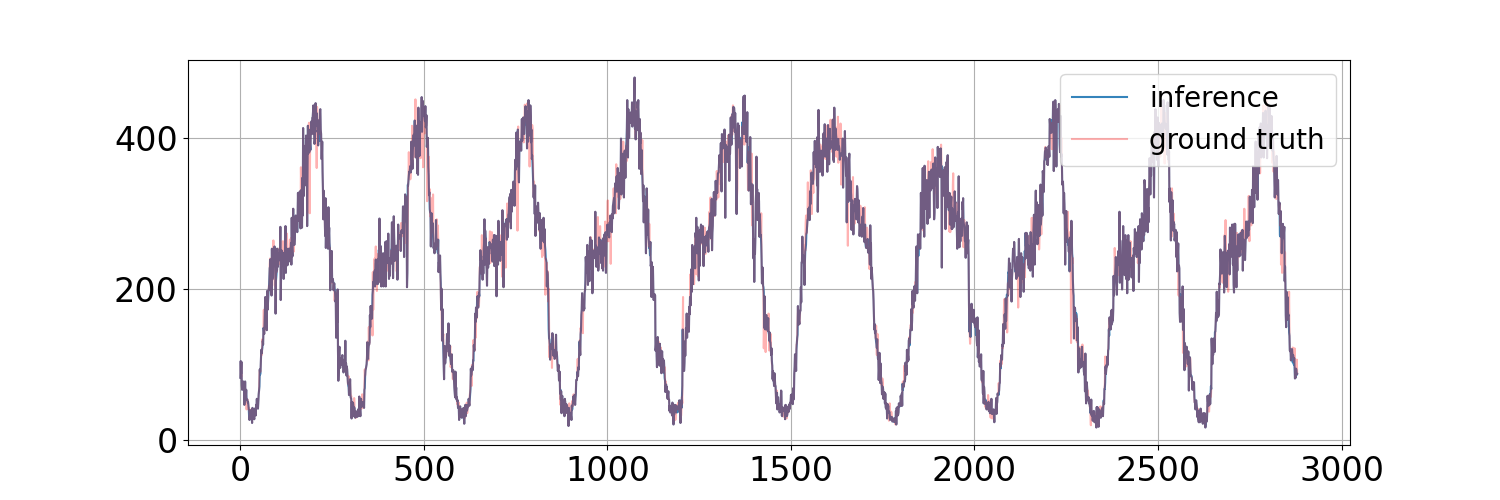}
        \caption{Point missing imputation on PEMSD8 on Node\#69.}
        \label{fig:sub2}
    \end{subfigure}
    \begin{subfigure}[b]{0.45\textwidth}
        \includegraphics[width=\textwidth]{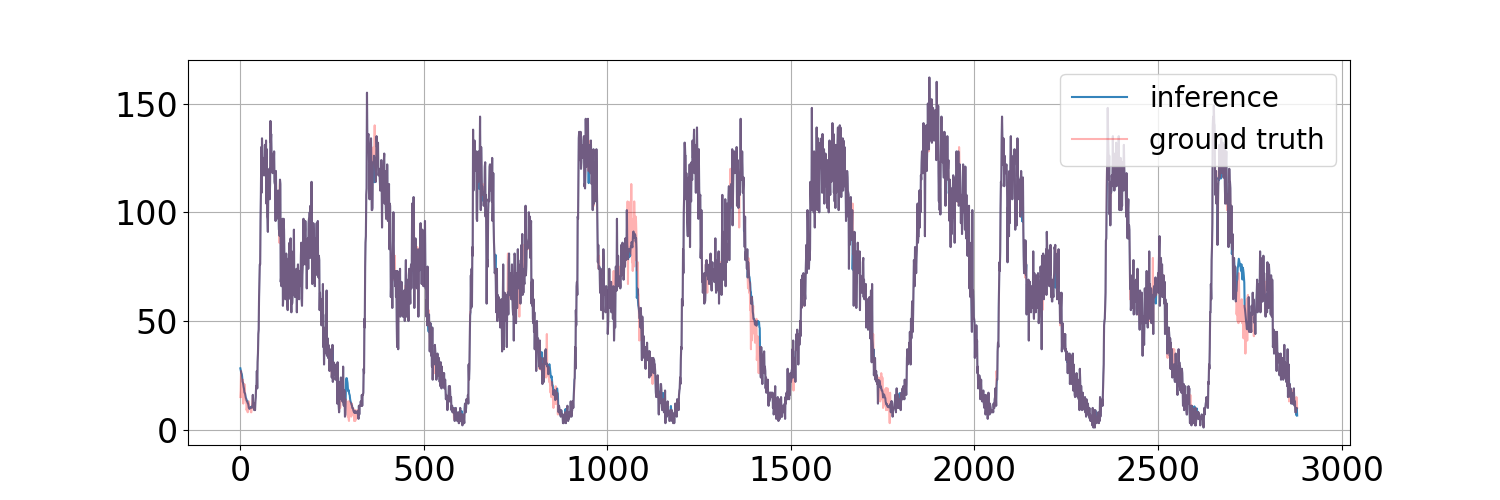}
        \caption{Block missing imputation on PEMSD8 on Node\#24.}
        \label{fig:sub3}
    \end{subfigure}
    \begin{subfigure}[b]{0.45\textwidth}
        \includegraphics[width=\textwidth]{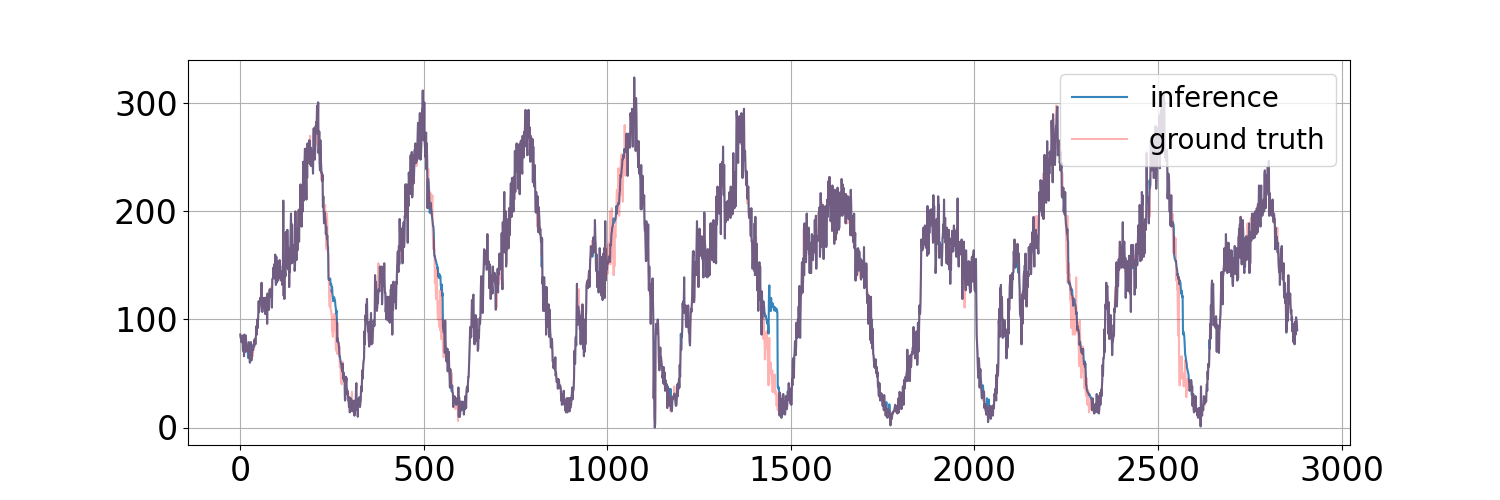}
        \caption{Block missing imputation on PEMSD8 on Node\#129.}
        \label{fig:sub4}
    \end{subfigure}
    \caption{Case Study of Efficiency on PEMSD8.}
    \label{fig_visualization}
\end{figure}

\section{Discussion of Semi-adaptive Dynamic Graph Structure Learning}
The semi-adaptive approach reveals how sampled attention approximates the cohesive factor through dynamic community representation to maintain spatial relationships. 
Additionally, a refraction vector helps align and reconstruct adjacency information, enabling more effective vector fusion.
Suppose that there is a matrix $\mathscr{A}$ in the space $\mathscr{S}$ that perfectly interprets the adjacency relationship. 
Its approximate low-rank factorization result is $\mathcal{E}^c(\mathcal{E}^e)^\top$ (we call they the cohesion and extroversion factors). 
The adaptive method is to learn $E_1~\text{and}~E_2$ through spatial features and make them close to $\mathcal{E}^c~\text{and}~\mathcal{E}^e$.
In our proposed semi-adaptive method, sampled attention approximates the cohesive factor in the low-rank factorization.
As described above, sampled attention reflects the distribution of the global influence of several essential nodes.
When these essential nodes can optimally propagate spatial features, they can be considered to represent a local community, respectively.
Then, spatial adjacency relationships can also be propagated through these community representatives.
At the same time, we cannot ignore the dynamics of sampled attention as a cohesive factor.
The community representatives extracted each time are necessarily disordered and different.
Consequently, we introduce a refraction vector $E^{ref}$ to refract the projected message into an externalization factor of aligned correlation. 
We dynamically guide the attention mechanism to reconstruct and refine the adjacency matrix, resulting in a more accurate representation of the underlying structure.

\section{Complexity Analysis}
The computational complexity of our STAMImputer model is $\mathcal{O}(T^2D + N \log ND + NTD)$, where $T$, $N$, and $D$ denote the temporal, spatial, and feature dimensions, respectively. 
This overall complexity arises from three components: the Temporal Expert (MSAT) with complexity $\mathcal{O}(T^2D)$, the Spatial Expert (LrSGAT) with $\mathcal{O}(N \log ND)$, and the Observation Expert with $\mathcal{O}(NTD)$.
Besides, the complexity of wavelet disentangling transform and the dynamic graph structure learning layer is $\mathcal{O}(NT)$ and $\mathcal{O}(N^2\log N)$, respectively.

\section{Visualization and Case Study}
We provide several visualization examples in Figure~\ref{fig_visualization}.
In order to intuitively show the model performance, we visualize the inference results with different missing scenarios and the ground truth.
Experimental results demonstrate that STAMImputer effectively imputes missing traffic data across both time and space. 
Even in scenarios with blocks of missing data, the model maintains strong performance, supported by the integration of the MoE framework and LrSGAT.

\begin{figure}[t]
\centering
\includegraphics[width=\columnwidth]{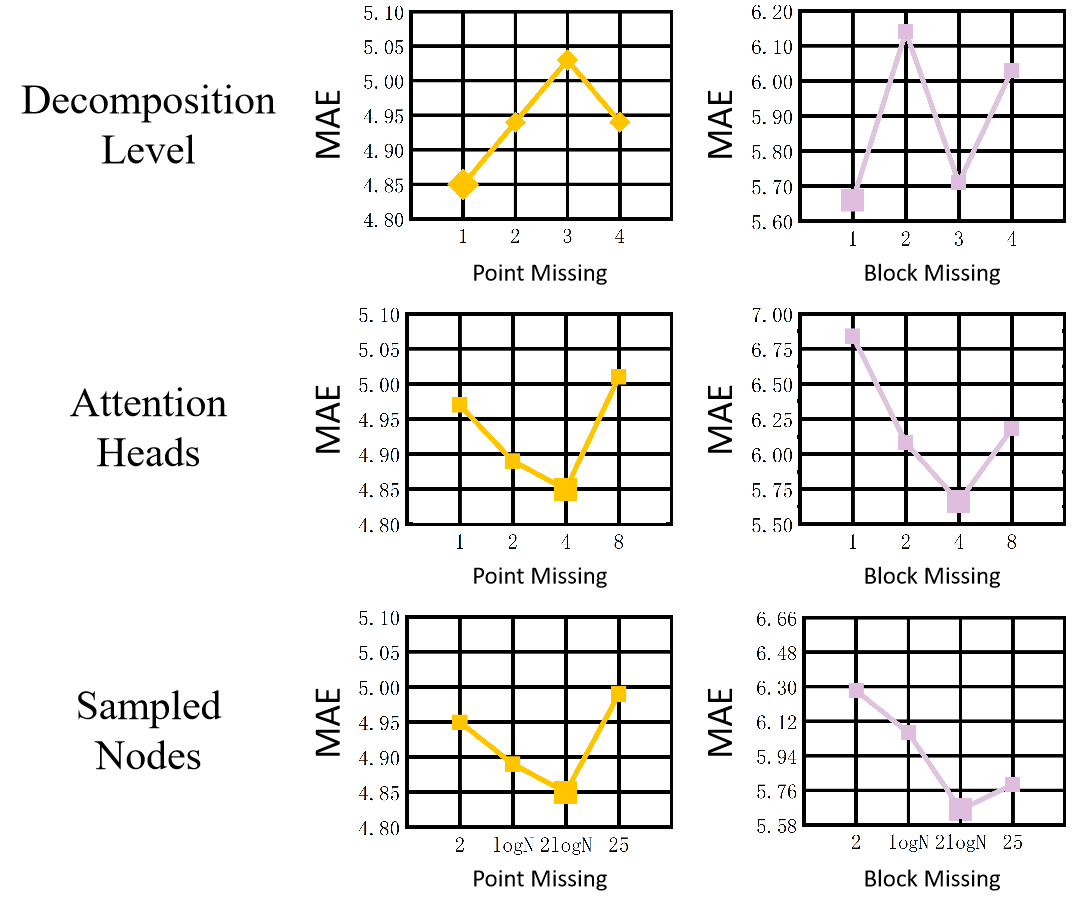}
\caption{
Case Study of Hyperparameters on NYC-Taxi benchmark.
}
\label{Fig_hyper}
\end{figure}

Figure~\ref{Fig_hyper} shows a sensitivity analysis of key hyperparameters—decomposition level, number of attention heads, and number of sampled nodes—under both point missing and block missing scenarios. 
The results show that changes in these hyperparameters lead to different MAE values, indicating that they have a noticeable effect on model performance.

\end{document}